\documentclass[runningheads]{llncs}
\usepackage{times}
\usepackage{amsmath}
\usepackage{amssymb}
\usepackage{graphicx}
\usepackage{algorithmic}
\usepackage{algorithm}
\usepackage{booktabs}
\usepackage{setspace}
\usepackage{hyperref}
\usepackage{cleveref}
\usepackage{listings}
\usepackage{esvect}
\usepackage{todonotes}
\usepackage{xspace}
\newcommand{\approach}{\textsc{Pyke}\xspace}
\newcommand{\Prob}{\ensuremath{\mathbb{P}}}
\usepackage{tabularx}
\usepackage{comment}
\usepackage{afterpage}
\usepackage[T1]{fontenc}
\usepackage[font=small,labelfont=bf,tableposition=top]{caption}
\usepackage[utf8]{inputenc}

\DeclareCaptionLabelFormat{andtable}{#1~#2  \&  \tablename~\thetable}

\begin{document}
\title{A Physical Embedding Model for Knowledge Graphs
\thanks{This work was supported by the German Federal Ministry of Transport and Digital Infrastructure project OPAL (GA: 19F2028A) as well as the H2020 Marie \text{Skłodowska}-Curie project KnowGraphs (GA no. 860801).}}

\titlerunning{A Physical Embedding Model for Knowledge Graphs}
\author{Caglar Demir\and
Axel-Cyrille Ngonga Ngomo}

\authorrunning{Demir and Ngonga Ngomo}
\institute{Paderborn University\\DICE  Research Group\\33098 Paderborn Germany \\
\email{{first.lastname\text{@}upb.de}}}
\maketitle
\begin{abstract}

Knowledge graph embedding methods learn continuous vector representations for entities in knowledge graphs and have been used successfully in a large number of applications.  We present a novel and scalable paradigm for the computation of knowledge graph embeddings, which we dub \approach. Our approach combines a  physical model based on Hooke's law and its inverse with ideas from simulated annealing to compute embeddings for knowledge graphs efficiently. We prove that \approach achieves a linear space complexity. While the time complexity for the initialization of our approach is quadratic, the time complexity of each of its iterations is linear in the size of the input knowledge graph. Hence, \approach's overall runtime is close to linear. Consequently, our approach easily scales up to knowledge graphs containing millions of triples. We evaluate our approach against six state-of-the-art embedding approaches on the DrugBank and DBpedia datasets in two series of experiments. The first series shows that the cluster purity achieved by \approach is up to 26\% (absolute) better than that of the state of art. In addition, \approach is more than 22 times faster than existing embedding solutions in the best case. The results of our second series of experiments show that \approach is up to 23\% (absolute) better than the state of art on the task of type prediction while maintaining its superior scalability. Our implementation and results are open-source and are available at {\url{http://github.com/dice-group/PYKE}}.
\end{abstract}

\section{Introduction}
The number and size of knowledge graphs (KGs) available on the Web and in companies grows steadily.\footnote{\url{https://lod-cloud.net/}} For example, more than 150 billion facts describing more than 3 billion things are available in the more than 10,000 knowledge graphs published on the Web as Linked Data.\footnote{\url{lodstats.aksw.org}} Knowledge graph embedding (KGE) approaches aim to map the entities contained in knowledge graphs to $n$-dimensional vectors \cite{kge_survey,rescal,DistMult}. Accordingly, they parallel word embeddings from the field of natural language processing \cite{mikolov2013distributed,pennington2014glove} and the improvement they brought about in various tasks (e.g., word analogy, question answering, named entity recognition and relation extraction). Applications of KGEs include collective machine learning, type prediction, link prediction, entity resolution, knowledge graph completion and question answering \cite{rescal,transe,nickel2016holographic,kge_survey,DistMult,rdf2vec}. In this work, we focus on type prediction.
We present a novel approach for KGE based on a physical model, which goes beyond the state of the art (see \cite{kge_survey} for a survey) w.r.t. both efficiency and effectiveness. Our approach, dubbed \approach, combines a \emph{physical model} (based on Hooke's law) 
with an optimization technique inspired by \emph{simulated annealing}. \approach scales to large KGs by achieving a linear space complexity while being close to linear in its time complexity on large KGs. We compare the performance of \approach with that of six state-of-the-art approaches---Word2Vec \cite{mikolov2013distributed}, ComplEx \cite{trouillon2016complex}, RESCAL \cite{rescal}, TransE \cite{transe}, DistMult \cite{DistMult} and Canonical Polyadic (CP) decomposition \cite{hitchcock1927expression}--- on two tasks, i.e., clustering and type prediction w.r.t. both runtime and prediction accuracy. Our results corroborate our formal analysis of \approach and suggest that our approach scales close to linearly with the size of the input graph w.r.t. its runtime. In addition to outperforming the state of the art w.r.t.  runtime, \approach also achieves better cluster purity and type prediction scores. 

The rest of this paper is structured as follows: after
providing a brief overview of related work in \Cref{relatedwork}, we present the mathematical framework underlying \approach in \Cref{preliminaries}. Thereafter, we
present \approach in \Cref{model}. 
\Cref{Complexity} presents the space and time complexity of \approach.
We report on the results of our experimental evaluation in \Cref{eval}. Finally, we conclude with a discussion and an outlook on future work in \Cref{conclusion}.

\section{Related Work}
\label{relatedwork}
A large number of KGE approaches have been developed to address tasks such as link prediction, graph completion and question answering \cite{KGE_QA,transR,nickel2016holographic,rescal,trouillon2016complex} in the recent past. In the following, we give a brief overview of some of these approaches. More details can be found in the survey at \cite{kge_survey}. RESCAL \cite{rescal} is based on computing a three-way factorization of an adjacency tensor representing the input KG. The adjacency tensor is decomposed into a product of a core tensor and embedding matrices.RESCAL captures rich interactions in the input KG but is limited in its scalability. HolE \cite{nickel2016holographic} uses circular correlation as its compositional operator. Holographic embeddings of knowledge graphs yield state-of-the-art results on link prediction task while keeping the memory complexity lower than RESCAL and TransR \cite{transR}.
ComplEx \cite{trouillon2016complex} is a KGE model based on latent factorization, wherein complex valued embeddings are utilized to handle a large variety of binary relations including symmetric and antisymmetric relations. 

Energy-based KGE models \cite{semantic_matching_energy_function,transe,bordes2011learning}
yield competitive performances on link prediction, graph completion and entity resolution. SE \cite{bordes2011learning} proposes to learn one low-dimensional vector ($\mathbb{R}^k$) for each entity and two matrices ($R_1 \in \mathbb{R}^{k \times k}$, $R_2 \in \mathbb{R}^{k \times k}$) for each relation. Hence, for a given triple ($h, r, t$), SE aims to minimize the $L_1$ distance, i.e., $f_r(h,t) = ||R_1 h - R_2 t||$. The approach in  \cite{semantic_matching_energy_function} embeds entities and relations into the same embedding space and suggests to capture correlations between entities and relations by using multiple matrix products. 
TransE \cite{transe} is a scalable energy-based KGE model wherein a relation $r$ between entities $h$ and $t$ corresponds to a translation of their embeddings, i.e., $h + r \approx t$ provided that $(h, r, t)$ exists in the KG. TransE outperforms state-of-the-art models in the link prediction task on several benchmark KG datasets while being able to deal with KGs containing up to 17 million facts. DistMult \cite{DistMult} proposes to generalize neural-embedding models under an unified learning framework, wherein relations are bi-linear or linear mapping function between embeddings of entities.

With \approach, we propose a different take to generating embeddings by combining a physical model with simulated annealing. Our evaluation suggests that this simulation-based approach to generating embeddings scales well (i.e., linearly in the size of the KG) while outperforming the state of the art in the type prediction and clustering quality tasks \cite{KG_w_entitiy_descrip,Wang2017CommunityPN}.

\section{Preliminaries and Notation}
	\label{preliminaries}
    In this section, we present the core notation and terminology used throughout this paper. The symbols we use and their meaning are summarized in \Cref{tab:notation}.

	\subsection{Knowledge Graph} 
	In this work, we compute embeddings for RDF KGs. Let $\mathcal{R}$ be the set of all RDF resources, $\mathcal{B}$ be the set of all RDF blank nodes, $\mathcal{P} \subseteq \mathcal{R}$ be the set of all properties and $\mathcal{L}$ denote the set of all RDF literals. An RDF KG $\mathcal{G}$ is a set of RDF triples $(s,p,o)$ where $s \in \mathcal{R} \cup \mathcal{B}$, $p \in \mathcal{P}$ and $o \in \mathcal{R} \cup  \mathcal{B} \cup \mathcal{L}$.
	We aim to compute embeddings for resources and blank nodes. Hence, we define the \emph{vocabulary} of an RDF knowledge graph $\mathcal{G}$ as $\mathcal{V} = \{x: x \in \mathcal{R} \cup \mathcal{P} \cup \mathcal{B} \wedge \exists(s, p, o) \in \mathcal{G}: x \in \{s,p,o\}\}$. Essentially, $\mathcal{V}$ stands for all the URIs and blank nodes found in $\mathcal{G}$. Finally, we define the \emph{subjects with type information} of $\mathcal{G}$ as $\mathcal{S} = \{x: x \in  \mathcal{R} \setminus \mathcal{P} \wedge (x,\texttt{rdf:type},o) \in \mathcal{G}\}$, where \texttt{rdf:type} stands for the instantiation relation in RDF.
	
\begin{table}[htb]
        \centering
        \caption{Overview of our notation}\label{params}
        \label{tab:notation}
            \begin{tabularx}{1\linewidth}{@{}p{.13\columnwidth}p{.8\columnwidth}@{}} 
            \toprule
            \textbf{Notation} & \textbf{Description} \\ \midrule
		$\mathcal{G}$ & An RDF knowledge graph \\
		$\mathcal{R},\mathcal{P},\mathcal{B},\mathcal{L}$ & Set of all RDF resources, predicates, blank nodes and literals respectively\\
		$\mathcal{S}$ & Set of all RDF subjects with type information \\
		$\mathcal{V}$ & Vocabulary of $\mathcal{G}$\\
		$\sigma$ & Similarity function on $\mathcal{V}$\\
		$\overrightarrow{x}_t$ & Embedding of $x$ at time $t$ \\
		$F_a \, , F_r$ & Attractive and repulsive forces, respectively \\
		${K}$ & Threshold for positive and negative examples\\
		$P$ & Function mapping each $x \in \mathcal{V}$ to a set of attracting elements of $\mathcal{V}$\\
		$N$ & Function mapping each $x \in \mathcal{V}$ to a set of repulsive elements of $\mathcal{V}$\\
		\Prob & Probability \\
		$\omega$ & Repulsive constant \\
		$\mathcal{E}$ & System energy \\
		$\epsilon$ & Upper bound on alteration of locations of $x \in \mathcal{V}$ across two iterations \\
		$\Delta e$ & Energy release \\
        \bottomrule
        \end{tabularx}
\end{table}
	
	\subsection{Hooke's Law} 
	Hooke's law describes the relation between a deforming force on a spring and the magnitude of the deformation within the elastic regime of said spring. The increase of a deforming force on the spring is linearly related to the increase of the magnitude of the corresponding deformation. In equation form, Hooke's law can be expressed as follows:
	\begin{equation}
	\label{hooke}
	F = - k \, \Delta
	\end{equation}
	where $F$ is the deforming force, $\Delta$ is the magnitude of deformation and $k$ is the spring constant. Let us assume two points of unit mass located at $x$ and $y$ respectively. We assume that the two points are connected by an ideal spring with a spring constant $k$, an infinite elastic regime and an initial length of 0. Then, the force they are subjected to has a magnitude of $k||x - y||$. Note that the magnitude of this force grows with the distance between the two mass points. 
	
	The inverse of Hooke's law, where 
    \begin{equation}
    \label{inverse_hooke}
        F_{\text{inverse}} = - \frac{k}{\Delta}
    \end{equation}
	has the opposite behavior. It becomes weaker with the distance between the two mass points it connects.
	 
	\subsection{Positive Pointwise Mutual Information} 
	The Positive Pointwise Mutual Information (PPMI) is a means to capture the strength of the association between two events (e.g., appearing in a triple of a KG). Let $a$ and $b$ be two events. Let $\Prob(a, b)$ stand for the joint probability of $a$ and $b$, $\Prob(a)$ for the probability of $a$ and $\Prob(b)$ for the probability of $b$. Then, $PPMI(a, b)$ is defined as
	\begin{equation}
		PPMI(a, b) = \max\left(0, \log\frac{\Prob(a, b)}{\Prob(a)\Prob(b)}\right),
	\end{equation}
 The equation truncates all negative values to 0 as measuring the strength of dissociation between events accurately demands very large sample sizes, which are empirically seldom available.
	
\section{\approach}
\label{model}

In this section, we introduce our novel KGE approach dubbed \approach (a \underline{p}h\underline{y}sical model for \underline{k}nowledge graph \underline{e}mbeddings). \Cref{premise} presents the intuition behind our model. In \Cref{framework}, we give an overview of the \approach framework, starting from processing the input KG to learning embeddings for the input in a vector space with a predefined number of dimensions. The workflow of our model is further elucidated using the running example shown in \Cref{input_rdf_graph}. 
	
\subsection{Intuition}
	\label{premise}
	\approach is an iterative approach that aims to represent each element $x$ of the vocabulary $\mathcal{V}$ of an input KG $\mathcal{G}$ as an embedding (i.e., a vector) in the $n$-dimensional space $\mathbb{R}^n$. Our approach begins by assuming that each element of $\mathcal{V}$ is mapped to a single point (i.e., its \emph{embedding}) of unit mass whose location can be expressed via an $n$-dimensional vector in $\mathbb{R}^n$ according to an initial (e.g., random) distribution at iteration $t=0$. In the following, we will use $\overrightarrow{x}_t$ to denote the embedding of $x \in \mathcal{V}$ at iteration $t$. We also assume a similarity function $\sigma: \mathcal{V} \times \mathcal{V} \rightarrow [0, \infty)$ (e.g., a PPMI-based similarity) over $\mathcal{V}$ to be given. Simply put, our goal is to improve this initial distribution iteratively over a predefined maximal number of iterations (denoted $T$) by ensuring that 
	\begin{enumerate}
		\item the embeddings of similar elements of $\mathcal{V}$ are close to each other while 
		\item the embeddings of dissimilar elements of $\mathcal{V}$ are distant from each other.
	\end{enumerate}

    Let $d: \mathbb{R}^n \times \mathbb{R}^n \rightarrow \mathbb{R}^+$ be the distance (e.g., the Euclidean distance) between two embeddings in $\mathbb{R}^n$. According to our goal definition, a good iterative embedding approach should have the following characteristics:
	\begin{enumerate}
		\item[$C_1$:] If $\sigma(x, y) > 0$, then $d(\overrightarrow{x}_t, \overrightarrow{y}_t) \leq d(\overrightarrow{x}_{t-1}, \overrightarrow{y}_{t-1})$. This means that the embeddings of similar terms should become more similar with the number of iterations. The same holds the other way around:
		 \item[$C_2$:] If $\sigma(x, y) = 0$, then $d(\overrightarrow{x}_{t}, \overrightarrow{y}_{t}) \geq d(\overrightarrow{x}_{t-1}, \overrightarrow{y}_{t-1})$.
	\end{enumerate}
	We translate $C_1$ into our model as follows: If $x$ and $y$ are similar (i.e., if $\sigma(x, y) > 0$), then a force $F_a(\overrightarrow{x}_t, \overrightarrow{y}_t)$ of attraction must exist between the masses which stand for $x$ and $y$ at any time $t$. $F_a(\overrightarrow{x}_t, \overrightarrow{y}_t)$ must be proportional to $d(\overrightarrow{x}_t, \overrightarrow{y}_t)$, i.e., the attraction between must grow with the distance between $(\overrightarrow{x}_t$ and $\overrightarrow{y}_t)$. These conditions are fulfilled by setting the following force of attraction between the two masses:
	\begin{equation}
		||F_a(\overrightarrow{x}_t, \overrightarrow{y}_t)|| = \sigma(x,y) \times d(\overrightarrow{x}_t, \overrightarrow{y}_t).
	\end{equation}  
	From the perspective of a physical model, this is equivalent to placing a spring with a spring constant of $\sigma(x,y)$ between the unit masses which stand for $x$ and $y$. At time $t$, these masses are hence accelerated towards each other with a total acceleration proportional to $||F_a(\overrightarrow{x}_t, \overrightarrow{y}_t)||$. 
	
	The translation of $C_2$ into a physical model is as follows: If $x$ and $y$ are not similar (i.e., if $\sigma(x, y) = 0$), we assume that they are dissimilar. Correspondingly, their embeddings should diverge with time. The magnitude of the repulsive force between the two masses representing $x$ and $y$ should be strong if the masses are close to each other and should diminish with the distance between the two masses. We can fulfill this condition by setting the following repulsive force between the two masses:
	\begin{equation}
	||F_r(\overrightarrow{x}_t, \overrightarrow{y}_t)|| = - \frac{ \omega }{d(\overrightarrow{x}_t, \overrightarrow{y}_t)},
	\end{equation}  
	where $\omega > 0$ denotes a constant, which we dub the repulsive constant. At iteration $t$, the embeddings of dissimilar terms are hence accelerated away from each other with a total acceleration proportional to $||F_r(\overrightarrow{x}_t, \overrightarrow{y}_t)||$. This is the inverse of Hooke's law, where the magnitude of the repulsive force between the mass points which stand for two dissimilar terms decreases with the distance between the two mass points.  
	
	Based on these intuitions, we can now formulate the goal of \approach formally: We aim to find embeddings for all elements of $\mathcal{V}$ which minimize the total distance between similar elements and maximize the total distance between dissimilar elements. Let $P: \mathcal{V} \rightarrow 2^\mathcal{V}$ be a function which maps each element of $\mathcal{V}$ to the subset of $\mathcal{V}$ it is similar to. Analogously, let $N: \mathcal{V} \rightarrow 2^\mathcal{V}$ map each element of $\mathcal{V}$ to the subset of $\mathcal{V}$ it is dissimilar to. \approach aims to optimize the following objective function:
	
	\begin{equation}
		J(\mathcal{V}) = \left(\sum\limits_{x \in \mathcal{V}} \sum\limits_{y \in P(x)} d(\overrightarrow{x}, \overrightarrow{y})\right) - \left(\sum\limits_{x \in \mathcal{V}} \sum\limits_{y \in N(x)} d(\overrightarrow{x}, \overrightarrow{y}) \right).
	\end{equation}
	
\subsection{Approach}
	\label{framework}
\begin{figure}[htb]
    \centering
        \includegraphics[width=.8\textwidth]{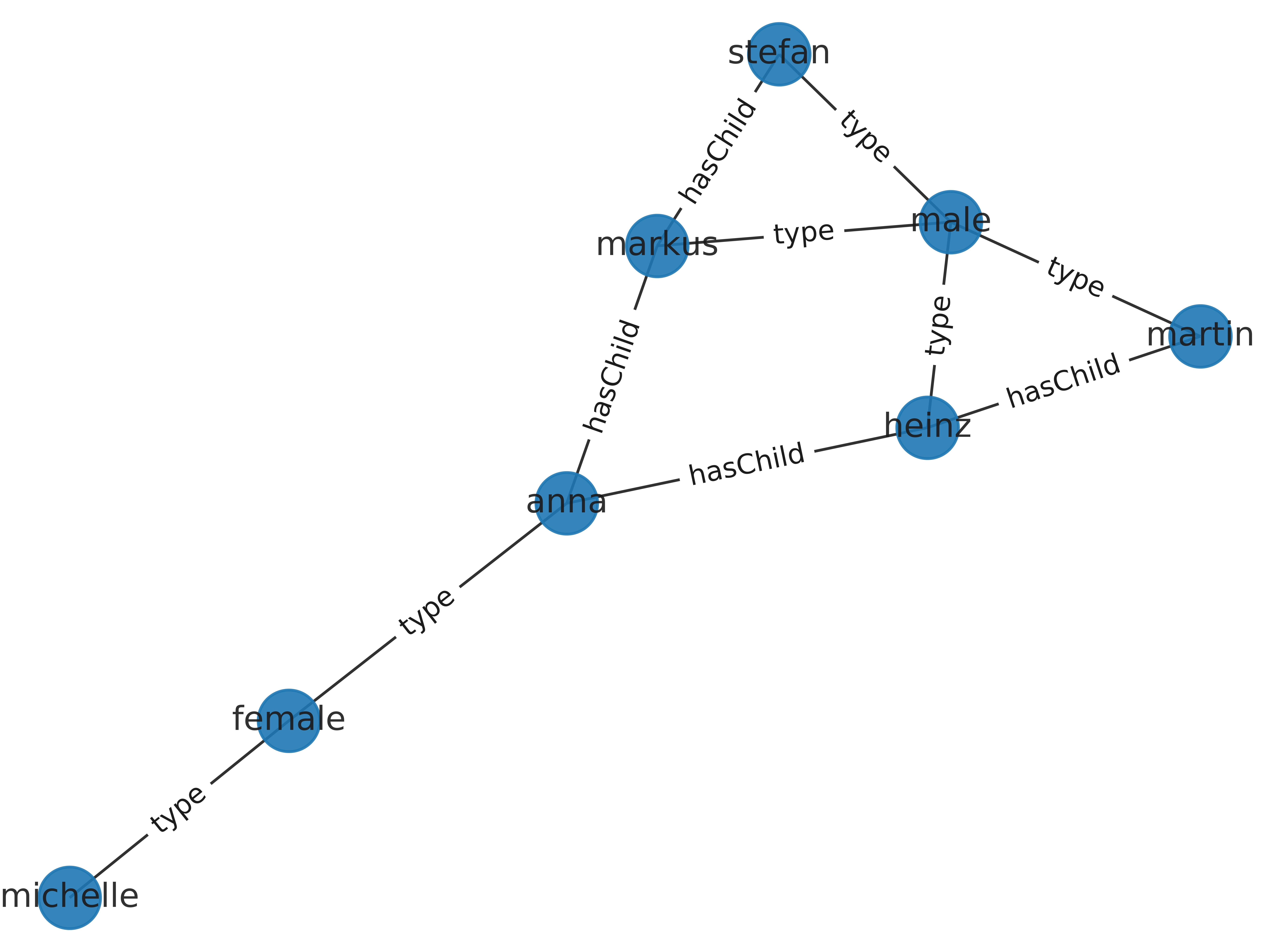}\caption{Example RDF graph}
	    \label{input_rdf_graph}
    \end{figure}

	\approach implements the intuition described above as follows: Given an  input KG $\mathcal{G}$, \approach first constructs a symmetric similarity matrix $\mathcal{A}$ of dimensions $|\mathcal{V}| \times |\mathcal{V}|$. We will use $a_{x,y}$ to denotes the similarity coefficient between $x \in \mathcal{V}$ and $y \in \mathcal{V}$ stored in $\mathcal{A}$.
	\approach truncates this matrix to (1) reduce the effect of oversampling and (2) accelerate subsequent computations. The initial embeddings of all $x \in \mathcal{V}$ in $\mathbb{R}^n$ are then determined. Subsequently, \approach uses the physical model described above to improve the embeddings iteratively. The iteration is ran at most $T$ times or until the objective function $J(\mathcal{V})$ stops decreasing. In the following, we explain each of the steps of the approach in detail. We use the RDF graph shown in \Cref{input_rdf_graph} as a running example.\footnote{This example is provided as an example in the DL-Learner framework at \url{http://dl-learner.org}.}
	
\begin{figure}[htb]
    \centering
    \includegraphics[width=.7\textwidth]{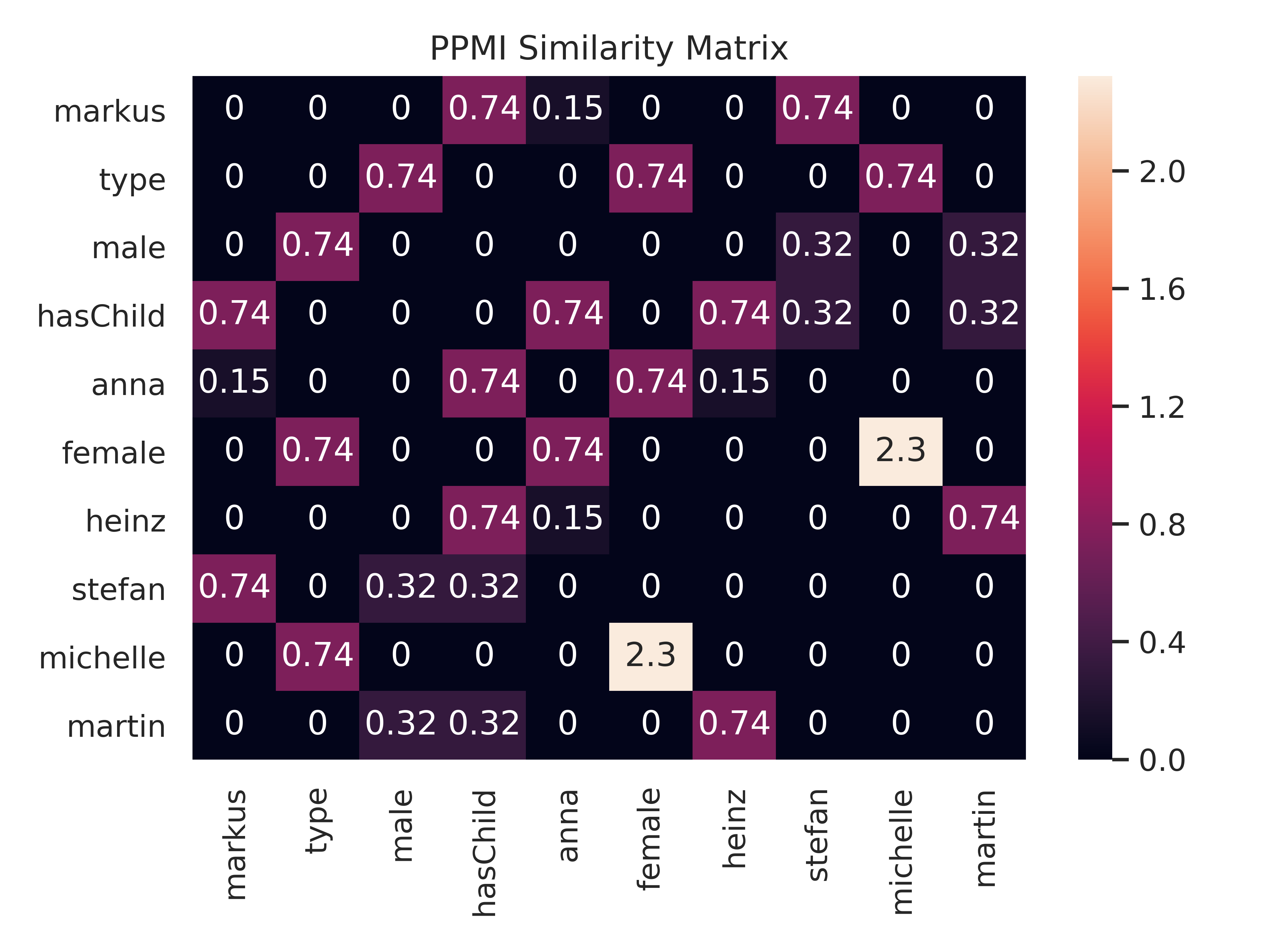}
    \caption{PPMI similarity matrix of resources in the RDF graph shown in Figure 1}
	\label{ppmi_matrix}
\end{figure}

	\subsubsection{Building the similarity matrix.}
	For any two elements $x, y \in \mathcal{V}$, we set $a_{x, y} = \sigma(x, y) = PPMI(x, y)$ in our current implementation. We compute the probabilities $\Prob(x)$, $\Prob(y)$ and $\Prob(x, y)$ as follows:
	\begin{equation}
		\Prob(x) = \frac{|\{(s, p, o) \in \mathcal{G}: x \in \{s, p, o\}\}|}{|\{(s, p, o) \in \mathcal{G}\}|}.
	\end{equation} 
	Similarly, 
	\begin{equation}
		\Prob(y) = \frac{|\{(s, p, o) \in \mathcal{G}: y \in \{s, p, o\}\}|}{|\{(s, p, o) \in \mathcal{G}\}|}.
	\end{equation}
	Finally, 
	\begin{equation}
		\Prob(x,y) = \frac{|\{(s, p, o) \in \mathcal{G}: \{x, y\} \subseteq \{s, p, o\}\}|}{|\{(s, p, o) \in \mathcal{G}\}|}.
	\end{equation}
	
    For our running example (see \Cref{input_rdf_graph}), \approach constructs the similarity matrix shown in \Cref{ppmi_matrix}. Note that our framework can be combined with any similarity function $\sigma$. Exploring other similarity function is out the scope of this paper but will be at the center of future works.
    
	\subsubsection{Computing $P$ and $N$.} To avoid oversampling positive or negative examples, we only use a portion of $\mathcal{A}$ for the subsequent optimization of our objective function.  For each $x \in \mathcal{V}$, we begin by computing $P(x)$ by selecting $K$ resources which are most similar to $x$. Note that if less than $K$ resources have a non-zero similarity to $x$, then $P(x)$ contains exactly the set of resources with a non-zero similarity to $x$. Thereafter, we sample $K$ elements $y$ of $\mathcal{V}$ with $a_{x, y} = 0$ randomly. We call this set $N(x)$. For all $y \in N(x)$, we set $a_{x,y}$ to $-\omega$, where $\omega$ is our repulsive constant. The values of $a_{x,y}$ for $y \in P(x)$ are preserved. All other values are set to 0. After carrying out this process for all $x \in \mathcal{V}$, each row of $\mathcal{A}$ now contains exactly $2K$ non-zero entries provided that each $x \in \mathcal{V}$ has at least $K$ resources with non-zero similarity. Given that $K << |\mathcal{V}|$, $\mathcal{A}$ is now sparse and can be stored accordingly.\footnote{We use $\mathcal{A}$ for the sake of explanation. For practical applications, this step can be implemented using priority queues, hence making quadratic space complexity for storing $\mathcal{A}$ unnecessary.} The PPMI similarity matrix for our example graph is shown in \Cref{ppmi_matrix}.

    \subsubsection{Initializing the embeddings.} 
\begin{figure}[htb]
    \centering
            \includegraphics[width=.49\textwidth]{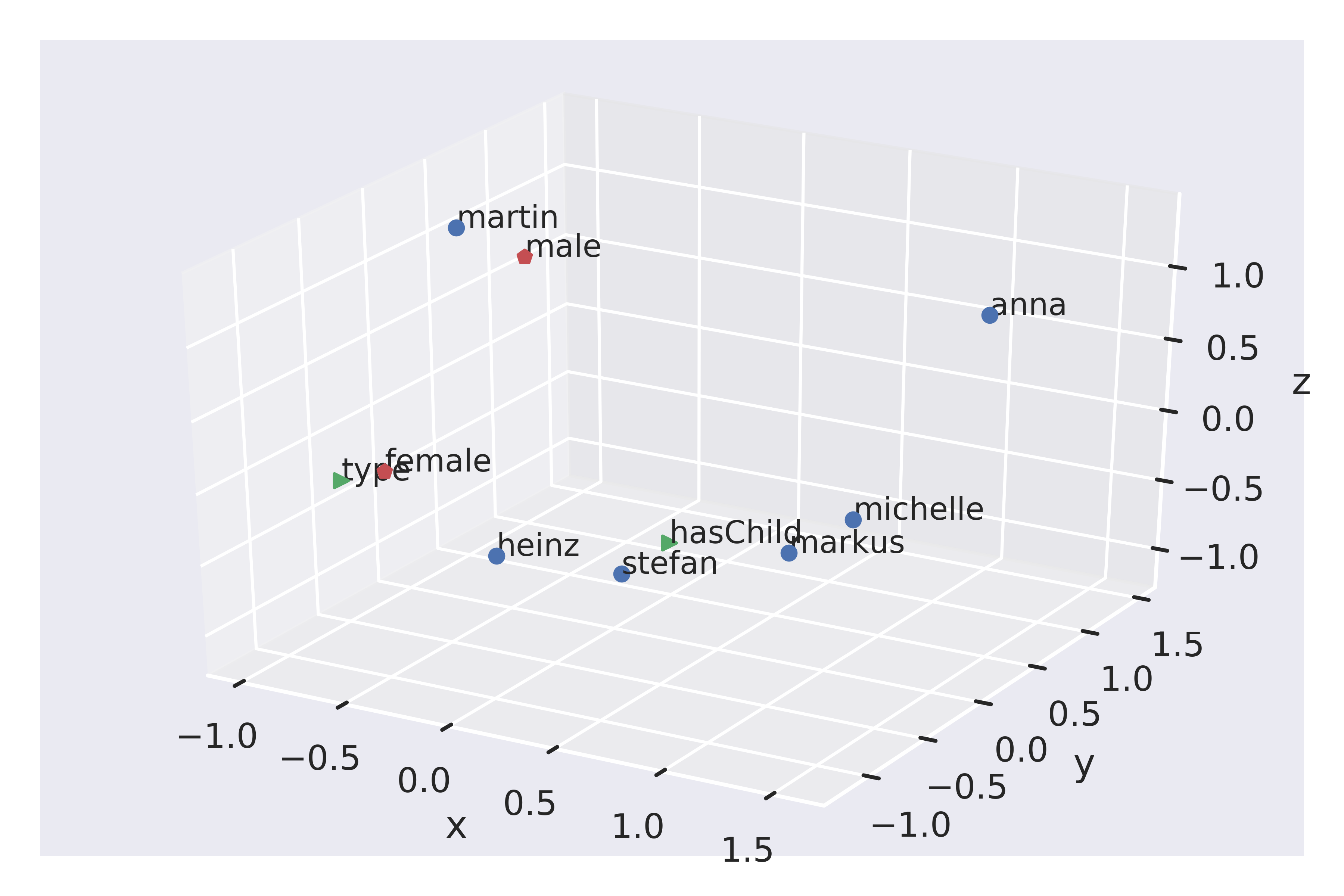}
            \hfill
            \includegraphics[width=.49\textwidth]{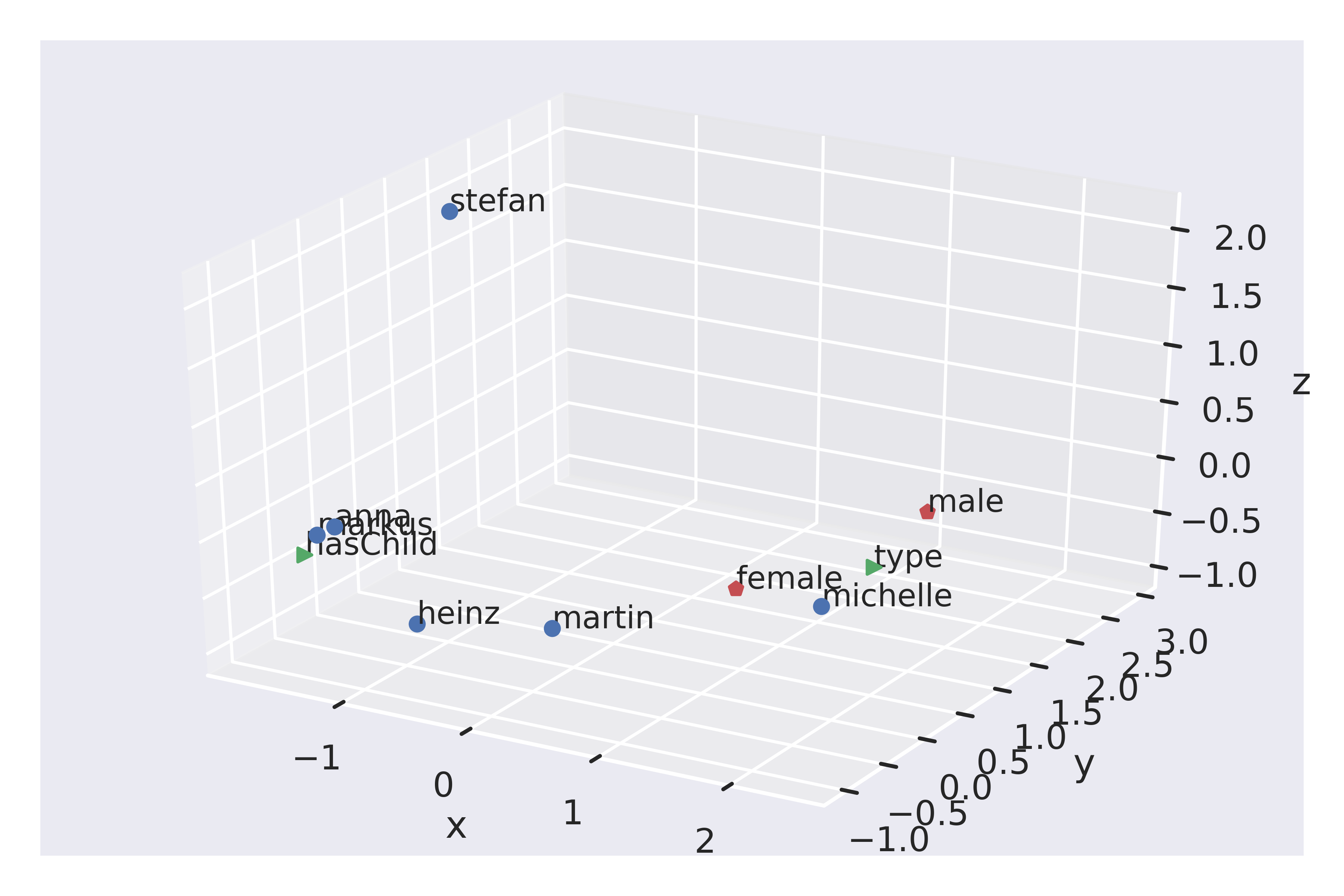}
	        \caption{PCA projection of 50-dimensional embeddings for our running example. Left are the randomly initialized embeddings. The figure on the right shows the 50-dimensional \approach embedding vectors for our running example after convergence. \approach was configured with $K=3$, $\omega=-0.3$, $\Delta e=0.06$ and $\epsilon=10^{-3}$.}
          \label{fig:embeddings}
        \end{figure}

Each $x \in \mathcal{V}$ is mapped to a single point $\overrightarrow{x}_t$ of unit mass in $\mathbb{R}^n$ at iteration $t=0$. As exploring sophisticated initialization techniques is out of the scope of this paper, the initial vector is set randomly.\footnote{Preliminary experiments suggest that applying a singular value decomposition on $\mathcal{A}$ and initializing the  embeddings with the latent representation of the elements of the vocabulary along the $n$ most salient eigenvectors has the potential of accelerating the convergence of our approach.} \Cref{fig:embeddings} shows a 3D projection of the initial embeddings for our running example (with $n = 50$).
	
	\subsubsection{Iteration.} This is the crux of our approach. In each iteration $t$, our approach assumes that the elements of $P(x)$ attract $x$ with a total force 
	\begin{equation}
	F_a(\overrightarrow{x}_t) = \sum\limits_{y \in P(x)} \sigma(x,y) \times (\overrightarrow{y}_t - \overrightarrow{x}_t).
	\end{equation}
	On the other hand, the elements of $N(x)$ repulse $x$ with a total force
	\begin{equation}
	F_r(\overrightarrow{x}_t) = - \sum\limits_{y \in N(x)}\frac{\omega} {(\overrightarrow{y}_t - \overrightarrow{x}_t)}.
	\end{equation}
	
	We assume that exactly one unit of time elapses between two iterations. The embedding of $x$ at iteration $t+1$  can now be calculated by displacing $\overrightarrow{x}_t$ proportionally to  $ (F_a(\overrightarrow{x}_t) + F_r(\overrightarrow{x}_t))$.However, implementing this model directly leads to a chaotic (i.e., non-converging) behavior in most cases. We enforce the convergence using an approach borrowed from simulated annealing, i.e., we reduce the total energy of the system by a constant factor $\Delta e$ after each iteration. By these means, we can ensure that  our approach always terminates, i.e., we can iterate until $J(\mathcal{V})$ does not decrease significantly or until a maximal number of iterations $T$ is reached. 

	\subsubsection{Implementation.}
	\Cref{alg:approach} shows the pseudocode of our approach. \approach updates the embeddings of vocabulary terms iteratively until one of the following two stopping criteria is satisfied: Either the upper bound on the iterations $T$ is met or a lower bound $\epsilon$ on the total change in the embeddings (i.e., $\sum\limits_{x \in \mathcal{V}} || \overrightarrow{x}_{t} - \overrightarrow{x}_{t-1}||$) is reached. A gradual reduction in the system energy $\mathcal{E}$ inherently guarantees the termination of the process of learning embeddings. A 3D projection of the resulting embedding for our running example is shown in \Cref{fig:embeddings}.
	
	\begin{algorithm}[h!tb]
		\caption{\approach}\label{alg:approach}
		\begin{algorithmic}[H]
			\REQUIRE $T$, $\mathcal{V}$, $K$, $\epsilon$, $\Delta e$, $\omega$, $n$ \\
			
			\STATE //initialize embeddings
			\FOR{ \textbf{each} $x$ in $\mathcal{V}$}
				\STATE $\overrightarrow{x}_0$ = random vector in $\mathbb{R}^n$;
			\ENDFOR
			
			\STATE //initialize similarity matrix
			\STATE $\mathcal{A}$ = new Matrix[$|\mathcal{V}|$][$|\mathcal{V}|$];
			
			\FOR{ \textbf{each} $x$ in $\mathcal{V}$}
				\FOR{ \textbf{each} $y$ in $\mathcal{V}$}
					\STATE $\mathcal{A}_{xy}$ = $PPMI(x, y)$;
				\ENDFOR
			\ENDFOR
			
			// perform positive and negative sampling
			\FOR{ \textbf{each} $x$ in $\mathcal{V}$}
				\STATE $P(x)$ = \text{getPositives}$( \mathcal{A}, {x},{K})$ ;
				\STATE $N(x)$ = \text{getNegatives}$(\mathcal{A}, {x},{K})$ ;
			\ENDFOR
	
			\STATE // iteration 
			\STATE $t=1$; 
			\STATE $\mathcal{E}=1$;
			
			\WHILE{$t < T$}
			
		    \FOR{ \textbf{each} $x$ in $\mathcal{V}$}
		    
    		    \STATE $F_a=\sum\limits_{y \in P(x)} \sigma(x,y)\times (\overrightarrow{y}_{t-1} - \overrightarrow{x}_{t-1})$;
                
                \STATE $F_r=- \sum\limits_{y \in N(x)} \frac{\omega} {\overrightarrow{y}_{t-1} - \overrightarrow{x}_{t-1}}$;
    			
			\STATE $\overrightarrow{x}_{t}=\overrightarrow{x}_{t-1}
			+ \mathcal{E} \times (F_a + F_r)$;
			\ENDFOR
			
			\STATE $\mathcal{E} = \mathcal{E} - \Delta e$;
			\IF{$ \sum\limits_{x \in \mathcal{V}}||\overrightarrow{x}_t - \overrightarrow{x}_{t-1}|| < \epsilon$} 
			\STATE{\textbf{break}}
			\ENDIF \\
			$t = t+1$;
			\ENDWHILE
			\RETURN Embeddings $\overrightarrow{x}_{t}$
		\end{algorithmic}
	\end{algorithm}

\section{Complexity Analysis}
\label{Complexity}

\subsection{Space complexity}
Let $m = |\mathcal{V}|$. We would need at most $\frac{m(m-1)}{2}$ entries to store $\mathcal{A}$, as the matrix is symmetric and we do not need to store its diagonal. However, there is actually no need to store $\mathcal{A}$. We can implement $P(x)$ as a priority queue of size $K$ in which the indexes of $K$ elements of $\mathcal{V}$ most similar to $x$ as well as their similarity to $x$ are stored. $N(x)$ can be implemented as a buffer of size $K$ which contains only indexes. Once $N(x)$ reaches its maximal size $K$, then new entries (i.e., $y$ with $PPMI(x, y)$) are added randomly. Hence, we need $O(Kn)$ space to store both $P$ and $N$. Note that $K << m$. The embeddings require exactly $2mn$ space as we store $\overrightarrow{x}_t$ and $\overrightarrow{x}_{t-1}$ for each $x \in \mathcal{V}$. The force vectors $F_a$ and $F_r$ each require a space of $n$. Hence, the space complexity of \approach lies clearly in $O(mn + Kn)$ and is hence linear w.r.t. the size of the input knowledge graph $\mathcal{G}$ when the number $n$ of dimensions of the embeddings and the number $K$ of positive and negative examples are fixed. 

\subsection{Time complexity}
Initializing the embeddings requires $mn$ operations. The initialization of $P$ and $N$ can also be carried out in linear time. Adding an element to $P$ and $N$ is carried out at most $m$ times. For each $x$, the addition of an element to $P(x)$ has a runtime of at most $K$. Adding elements to $N(x)$ is carried out in constant time, given that the addition is random. Hence the computation of $P(x)$ and $N(x)$ can be carried out in linear time w.r.t. $m$. This computation is carried out $m$ times, i.e., once for each $x$. Hence, the overall runtime of the initialization for \approach is on $O(m^2)$. Importantly, the update of the position of each $x$ can be carried out in $O(K)$, leading to each iteration having a time complexity of $O(mK)$. The total runtime complexity for the iterations is hence $O(mKT)$, which is linear in $m$. This result is of central importance for our subsequent empirical results, as the iterations make up the bulk of \approach's runtime. Hence, \approach's runtime should be close to linear in real settings.

\section{Evaluation}
\label{eval}
\subsection{Experimental Setup}
\label{subsec:expsetup}
The goal of our evaluation was to compare the quality of the embeddings generated by \approach with the state of the art. Given that there is no intrinsic measure for the quality of embeddings, we used two extrinsic evaluation scenarios. In the first scenario, we measured the type homogeneity of the embeddings generated by the KGE approaches we considered. We achieved this goal by using a scalable approximation of DBSCAN dubbed HDBSCAN \cite{hdbscan}. In our second evaluation scenario, we compared the performance of \approach on the type prediction task against that of 6 state-of-the-art algorithms. In both scenarios, we only considered embeddings of the subset $\mathcal{S}$ of $\mathcal{V}$ as done in previous works \cite{melo2016type,thoma2017towards}. We set $K=45$, $\Delta{}e = 0.0414$ and $\omega = 1.45557$ throughout our experiments. The values were computed using a Sobol Sequence optimizer \cite{saltelli2010variance}. All experiments were carried out on a single core of a server running Ubuntu 18.04 with $126$ GB RAM with 16 Intel(R) Xeon(R) CPU E5-2620 v4 @ 2.10GHz processors.

    We used six datasets (2 real, 4 synthetic)  throughout our experiments. An overview of the datasets used in our experiments is shown in \Cref{all_dataset_info}.
    Drugbank\footnote{\url{download.bio2rdf.org/\#/release/4/drugbank}} is a small-scale KG, whilst the DBpedia (version 2016-10) dataset is a large cross-domain dataset.\footnote{
    Note that we compile the DBpedia datasets by merging the dumps of \texttt{mapping-based objects}, \texttt{skos categories} and \texttt{instance types}  provided in the DBpedia download folder for version 2016-10 at \url{downloads.dbpedia.org/2016-10}.} The \emph{four synthetic datasets} were generated using the LUBM generator \cite{guo2005lubm} with 100, 200, 500 and 1000 universities. 

    \begin{table}[htb]
    	\centering
    	\caption{Overview of RDF datasets used in our experiments}
    	\label{all_dataset_info}
        \begin{tabular}{lrrrr}
    	    \toprule
        	Dataset & $|\mathcal{G}|$ & $|\mathcal{V}|$ & $|\mathcal{S}|$ & $|\mathcal{C}|$ \\
        	\midrule
        	Drugbank & 3,146,309 & 521,428  & 421,121 & 102 \\
        	DBpedia & 27,744,412 & 7,631,777 & 6,401,519 & 423 \\ \midrule
        	LUBM100 & 9,425,190 & 2,179,793 & 2,179,766 & 14\\
        	LUBM200 & 18,770,356 & 4,341,336 & 4,341,309 & 14 \\
    		LUBM500 & 46,922,188 & 10,847,210 & 10,847,183 & 14 \\
    		LUBM1000 & 93,927,191 & 21,715,108 & 21,715,081 & 14 \\
        	\bottomrule
        \end{tabular}
    \end{table}

We evaluated the homogeneity of embeddings by measuring the purity \cite{manning2010introduction} of the clusters generated by HDBSCAN \cite{hdbscan}. The original cluster purity equation assumes that each element of a cluster is mapped to exactly one class \cite{manning2010introduction}. Given that a single resource can have several types in a knowledge graph (e.g., \texttt{BarackObama} is a person, a politician, an author and a president in DBpedia), we extended the cluster purity equation as follows: Let $\mathcal{C} = \{c_1, c_2, \ldots \}$ be the set of all classes found in $\mathcal{G}$. Each $x \in \mathcal{S}$ was mapped to a binary type vector $type(x)$ of length $|\mathcal{C}|$. The ith entry of $type(x)$ was 1 iff $x$ was of type $c_i$. In all other cases, $c_i$ was set to 0. Based on these premises, we computed the purity of a clustering as follows: 
	
\begin{equation}
	\label{task_clustering_quality}
	\text{Purity}= \sum\limits_{l=1} ^{L}\frac{1}{|\zeta_l|^2} \sum\limits_{x \in \zeta_l} \sum\limits_{y \in \zeta_l} cos\Big(type(x) , type(y)\Big),
	\end{equation}
	where $\zeta_1 \ldots \zeta_L$ are the clusters computed by HDBSCAN. A high purity means that resources with similar type vectors (e.g., presidents who are also authors) are located close to each other in the embedding space, which is a wanted characteristic of a KGE. 

In our second evaluation, we performed a type prediction experiment in a manner akin to \cite{melo2016type,thoma2017towards}. For each resource $x \in \mathcal{S}$, we used the $\mu$ closest embeddings of $x$ to predict $x$'s type vector. We then compared the average of the types predicted with $x$'s known type vector using the cosine similarity:
	\begin{equation}
	\label{type_prediction}
	\text{prediction score}=  \frac{1}{|\mathcal{S}|} \sum\limits_{x \in \mathcal{S}} cos\Big(type(x) , \sum\limits_{y \in {\mu}nn(x)} type(y)\Big),
	\end{equation}
    where ${\mu}nn(x)$ stands for the $\mu$ neareast neighbors of $x$. 
We employed $\mu \in \{$1, 3, 5, 10, 15, 30, 50, 100$\}$ in our experiments. 

Preliminary experiments showed that performing the cluster purity and type prediction evaluations on embeddings of large knowledge graphs is prohibited by the long runtimes of the clustering algorithm. For instance, HDBSCAN did not terminate in 20 hours of computation when $|\mathcal{S}|>6 \times 10^6$. Consequently, we had to apply HDBSCAN on embeddings on the subset of $\mathcal{S}$ on DBpedia which contained resources of type \texttt{Person} or \texttt{Settlement}. The resulting subset of $\mathcal{S}$ on DBpedia consists of $428,289$ RDF resources. For the type prediction task, we sampled $10^5$  resources from $\mathcal{S}$ according to a random distribution and fixed them across the type prediction experiments for all KGE models. 
    
	\subsection{Results}

    \subsubsection{Cluster Purity Results.}
    \Cref{cluster_quality_results} displays the cluster purity results for all competing approaches. \approach achieves a cluster purity of 0.75 on Drugbank and clearly outperforms all other approaches. DBpedia turned out to be a more difficult dataset. Still, \approach was able to outperform all  state-of-the-art approaches by between 11\% and 26\% (absolute) on Drugbank and between 9\% and 23\% (absolute) on DBpedia.
    Note that in 3 cases, the implementations available were unable to complete the computation of embeddings within 24 hours. 
    
	\begin{table}[htb]
		\caption{Cluster purity results. The best results are marked in bold. Experiments marked with * did not terminate after 24 hours of computation.}
		\label{cluster_quality_results}
		\centering
			\begin{tabular}{p{2.5cm}p{2cm}p{2cm}}
				\toprule
				{\textbf{Approach}} &{\textbf{Drugbank}} &{\textbf{DBpedia}}\\
				\midrule
				\approach  & \textbf{0.75} & \textbf{0.57}\\
				Word2Vec   & 0.43  &  0.37 \\
				ComplEx    & 0.64  &  *  \\
				RESCAL     & *     &  *  \\
				TransE     & 0.60  &  0.48 \\
				CP         & 0.49  &  0.41 \\
				DistMult   & 0.49  &  0.34 \\
				\bottomrule
			\end{tabular}
	\end{table}
	
    \subsubsection{Type Prediction Results.}
\begin{figure}[htb]    
	 \centering
	 \includegraphics[width=.8\textwidth]{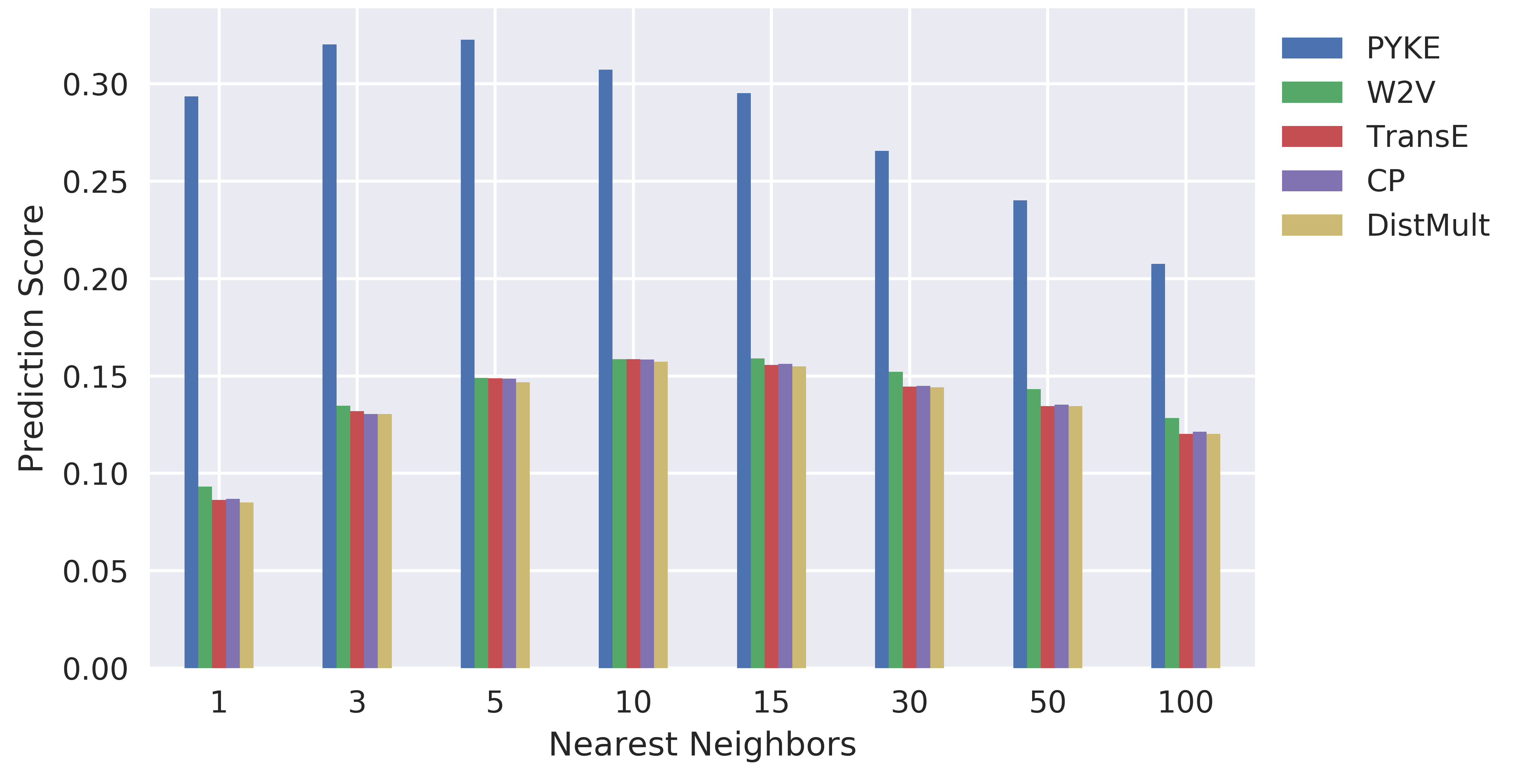}
	    \caption{Mean results on type prediction scores on $10^5$ randomly sampled entities of DBpedia}
	    \label{dbpedia_type_predictions}
	 \end{figure}

\begin{figure}[htb]
          \centering
	    \includegraphics[width=.8\textwidth]{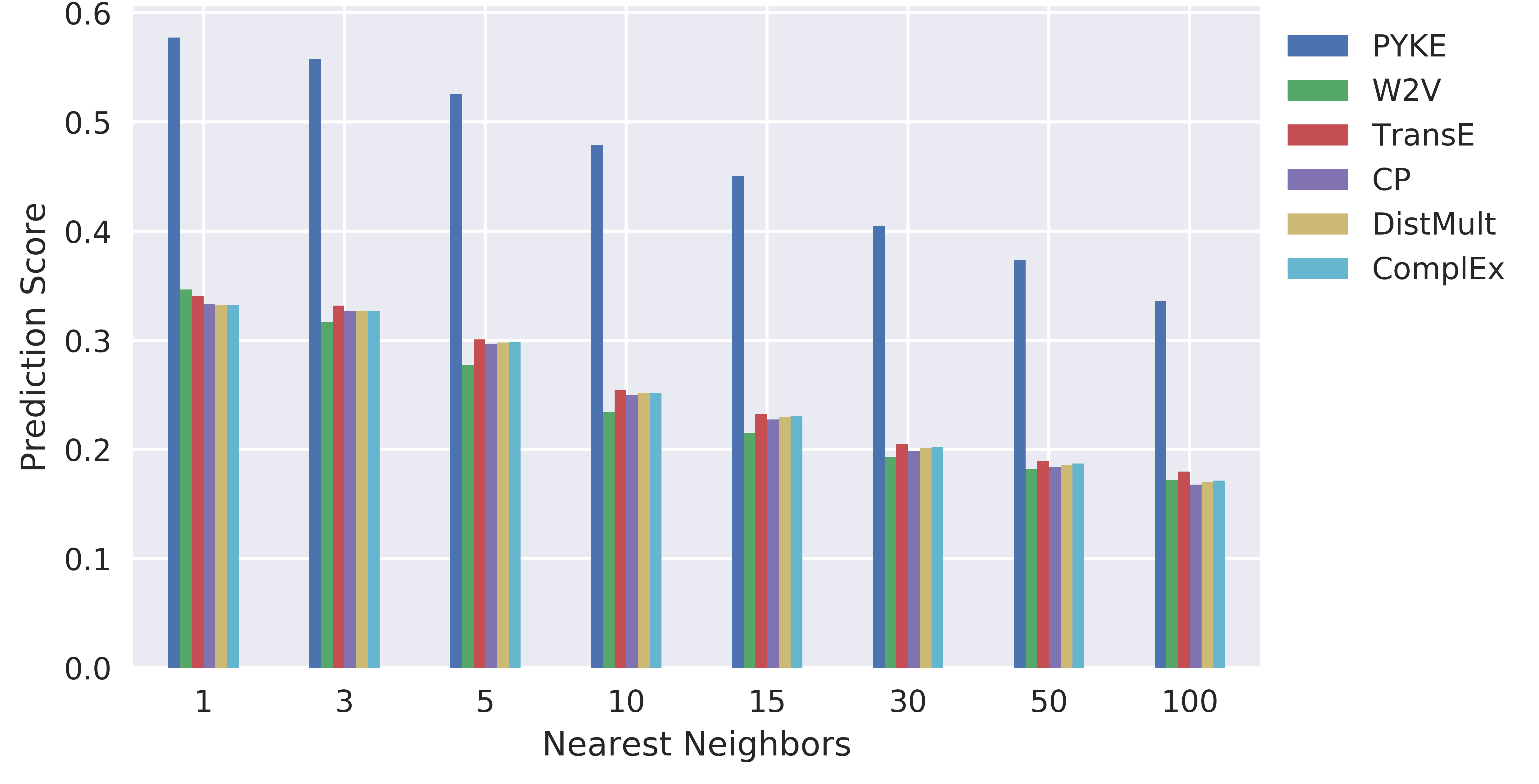}
	    \caption{Mean of type prediction scores on all entities of Drugbank}
	    \label{drugbank_type_predictions}
    \end{figure}

    \Cref{dbpedia_type_predictions} and \Cref{drugbank_type_predictions} show our type prediction results on the Drugbank and DBpedia datasets. \approach outperforms all state-of-the-art approaches across all experiments. In particular, it achieves a  margin of up to 22\% (absolute) on Drugbank and 23\% (absolute) on DBpedia. Like in the previous experiment, all KGE approaches perform worse on DBpedia, with prediction scores varying between $<0.1$ and $0.32$.

    \subsubsection{Runtime Results.}
\begin{figure}[htb]
		\centering
        \includegraphics[width=.6\textwidth]{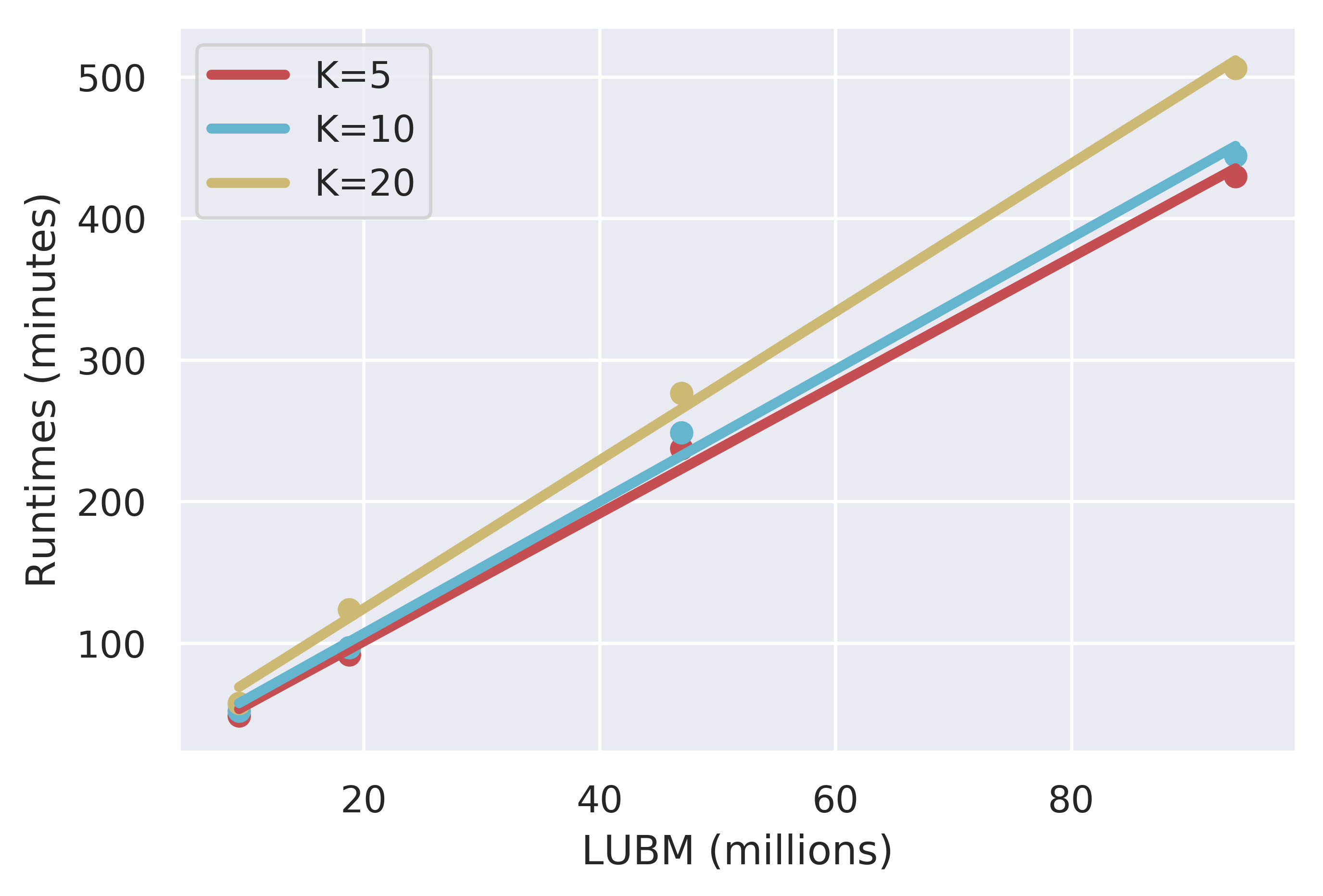}
        \caption{Runtime performances of \approach on synthetic KGs. Colored lines represent fitted linear regressions with fixed $K$ values of \approach.}
        \label{runtimes_on_lubm}
	\end{figure}

    \Cref{tab:runtimes} show runtime performances of all models on the two real benchmark datasets, while \Cref{runtimes_on_lubm} display the runtime of \approach on the synthetic LUBM datasets. Our results support our original hypothesis. The low space and time complexities of \approach mean that it runs efficiently: Our approach achieves runtimes of only 25 minutes on Drugbank and 309 minutes on DBpedia, while outperforming all other approaches by up to 14 hours in runtime. 
    
    In addition to evaluating the runtime of \approach on synthetic data, we were interested in determining its behaviour on datasets of growing sizes. We used LUBM datasets and computed a linear regression of the runtime using ordinary least squares (OLS). The runtime results for this experiment are shown in \Cref{runtimes_on_lubm}. The linear fit shown in \Cref{fitting_ols} achieves $R^2$ values beyond 0.99, which points to a clear linear fit between \approach's runtime and the size of the input dataset.

	\begin{table}[htb]
		\centering
		\begin{tabular}{cccc}
				\toprule
				{\textbf{K}}&{\textbf{Coefficient}} & {\textbf{Intercept}} & {\textbf{$R^2$}}\\
				\midrule
				5        & 4.52 & 10.74  &   0.997 \\
				10       & 4.65 & 13.64  &   0.996  \\
				20       & 5.23 & 19.59  &   0.997\\
				\bottomrule
			\end{tabular}
        \caption{Results of fitting OLS on runtimes.}
		\label{fitting_ols}
	\end{table}

    	\begin{table}[htb]
        \caption{Runtime performances (in minutes) of all competing approaches. All approaches were executed three times on each dataset. The reported results are the mean and standard deviation of the last two runs. The best results are marked in bold. Experiments marked with * did not terminate after 24 hours of computation.}
		\centering
		    \small
			\begin{tabular}{lcccccccccc}
				\toprule
				{\textbf{Approach}}& &{\textbf{Drugbank}} & & {\textbf{DBpedia}}\\
				\midrule
				\approach   && \textbf{25} $\pm$ 1 & & \textbf{309} $\pm$ 1     \\
				Word2Vec    && 41 $\pm 1$ & & 420 $\pm 1$     \\
				ComplEx     && 705 $\pm$ 1      & & * &      \\
				RESCAL      && *       & & * &      \\
				TransE      && 68 $\pm$ 1      & & 685 $\pm$ 1 \\
				CP          && 230 $\pm$ 1      & & 1154 $\pm$ 1     \\
				DistMult    && 210 $\pm$ 1      & & 1030 $\pm$ 1\\
				\bottomrule
			\end{tabular}
		\label{tab:runtimes}
	\end{table}

    We believe that the good performance of \approach stems from (1) its  sampling procedure and (2) its being akin to a physical simulation. Employing PPMI to quantify the similarity between resources seems to yield better sampling results than generating negative examples using the \emph{local closed word assumption} that underlies sampling procedures of all of competing state-of-the-art KG models. More importantly, positive and negative sampling occur in our approach per resource rather than per RDF triple. Therefore, \approach is able to leverage more from negative and positive sampling. By virtue of being akin to a physical simulation, \approach is able to run efficiently even when each resource $x$ is mapped to 45 attractive and 45 repulsive resources (see \Cref{tab:runtimes}) whilst all state-of-the-art KGE required more computation time.

\section{Conclusion}
\label{conclusion}
We presented \approach, a novel approach for the computation of embeddings on knowledge graphs. By virtue of being akin to a physical simulation, \approach retains a linear space complexity. This was proven through a complexity analysis of our approach. While the time complexity of the approach is quadratic due to the computation of $P$ and $N$, all other steps are linear in their runtime complexity. Hence, we expected our approach to behave closes to linearly. Our evaluation on LUBM datasets suggests that this is indeed the case and the runtime of  our approach grows close to linearly. This is an important result, as it means that our approach can be used on very large knowledge graphs and return results faster than popular algorithms such as Word2VEC and TransE. However, time efficiency is not all. 
Our results suggest that \approach outperforms state-of-the-art approaches in the two tasks of type prediction and clustering. Still, there is clearly a lack of normalized evaluation scenarios for knowledge graph embedding approaches. We shall hence develop such benchmarks in future works. 
Our results open a plethora of other research avenues. First, the current approach to compute similarity between entities/relations on KGs is based on the local similarity. Exploring other similarity means will be at the center of future works. In addition, using a better initialization for the embeddings should lead to faster convergence. Finally, one could use a stochastic approach (in the same vein as stochastic gradient descent) to further improve the runtime of \approach.

\bibliographystyle{splncs04}
\bibliography{references}

\end{document}